\documentclass[11pt]{article}
\usepackage[final]{rbs}
\usepackage[T1]{fontenc}
\usepackage{lmodern}
\usepackage{amsmath,amssymb,mathtools}
\usepackage{graphicx}
\usepackage{booktabs,multirow,tabularx}
\usepackage{xspace}
\usepackage{enumitem}

\usepackage{amsmath,amsfonts,bm}
\usepackage{xspace}
\usepackage{xcolor}

\def\eqref#1{equation~\ref{#1}}

\def\1{\bm{1}}

\DeclareMathAlphabet{\mathsfit}{\encodingdefault}{\sfdefault}{m}{sl}
\SetMathAlphabet{\mathsfit}{bold}{\encodingdefault}{\sfdefault}{bx}{n}

\usepackage{float}
\usepackage{subcaption}
\usepackage{wrapfig}
\usepackage{array}
\usepackage{hyperref}
\usepackage{url}
\hypersetup{colorlinks=true,linkcolor=rbsInk,citecolor=rbsInk,urlcolor=rbsGreen}

\newcommand{\method}{\textsc{WorldFlow}}

\title{WorldFlow: Joint Video and 3D Motion Prediction via Large-Scale Pretraining}

\author{%
\begin{minipage}{\dimexpr\textwidth-3\tabcolsep\relax}
\centering
{\bfseries Yuhao Lu\textsuperscript{1,*}, Jiongxu Chen\textsuperscript{1,*}, Hexiang Wei\textsuperscript{1}, Jiajin Wu\textsuperscript{1}, Juelin Li\textsuperscript{1}, Mingjian Liang\textsuperscript{1}} \\[0.3em]
{\bfseries Fan Yang\textsuperscript{1}, Zhuowei Liang\textsuperscript{1}, Zihao Liu\textsuperscript{1}, Zhennan Wang\textsuperscript{1}, Sheng Wang$^{1,\dagger}$, Lin Shao$^{1,2,\ddagger}$} \\[0.4em]
{\normalfont\small \textsuperscript{1}RoboScience \quad \textsuperscript{2}National University of Singapore}
\end{minipage}}

\begin{document}

\raggedbottom

\maketitle
\begingroup
\makeatletter
\renewcommand{\@makefntext}[1]{\noindent#1}
\makeatother
\renewcommand{\thefootnote}{}
\renewcommand{\theHfootnote}{rbscredits}
\begin{NoHyper}
\footnotetext{\textsuperscript{*} Equal contribution.\quad \textsuperscript{\ensuremath{\dagger}} Project leader.\quad \textsuperscript{\ensuremath{\ddagger}} Corresponding author.}
\end{NoHyper}
\endgroup

\begin{abstract}

Predicting future physical states stands as a central challenge in embodied intelligence. Existing video world models leave 3D dynamics implicit, action-based formulations remain tied to specific robot configurations, and 3D motion prediction models use flow-only supervision, providing a limited learning signal for physical dynamics. To circumvent these limitations, we introduce \method{}, a generative world model that jointly predicts future video and query-based 3D trajectories. By parameterizing 3D motion as continuous displacement tokens aligned with the video grid, a unified Mixture-of-Transformers (MoT) backbone jointly denoises both modalities under a rectified-flow objective. This formulation couples explicit geometric dynamics with visual evolution, effectively unlocking the physical commonsense embedded within pretrained video models. To empirically scale and validate this unified architecture, we train \method{} on a heterogeneous mixture of 1.44 million object-level samples spanning simulated and real-world domains. We further introduce WorldFlowBench, comprising 1,678
evaluation samples from 11 data
sources across simulated and real-world interactions,
including spatial reasoning tasks. Across all three
evaluation sets, WorldFlow achieves the lowest average
displacement error (ADE) and final displacement error
(FDE) among the evaluated learning-based methods.
Notably, on the challenging SpatialCog set, WorldFlow achieves an ADE of 0.060 m, a 54.2\% reduction compared with the strongest external baseline. It also reduces ADE and FDE by 55.9\% and 68.5\%, respectively, compared with its flow-only variant. Project page: \url{https://roboscience-ai.github.io/worldflow-page/}.

\end{abstract}

\begin{center}
\vspace{-0.25cm}
\includegraphics[width=0.98\linewidth]{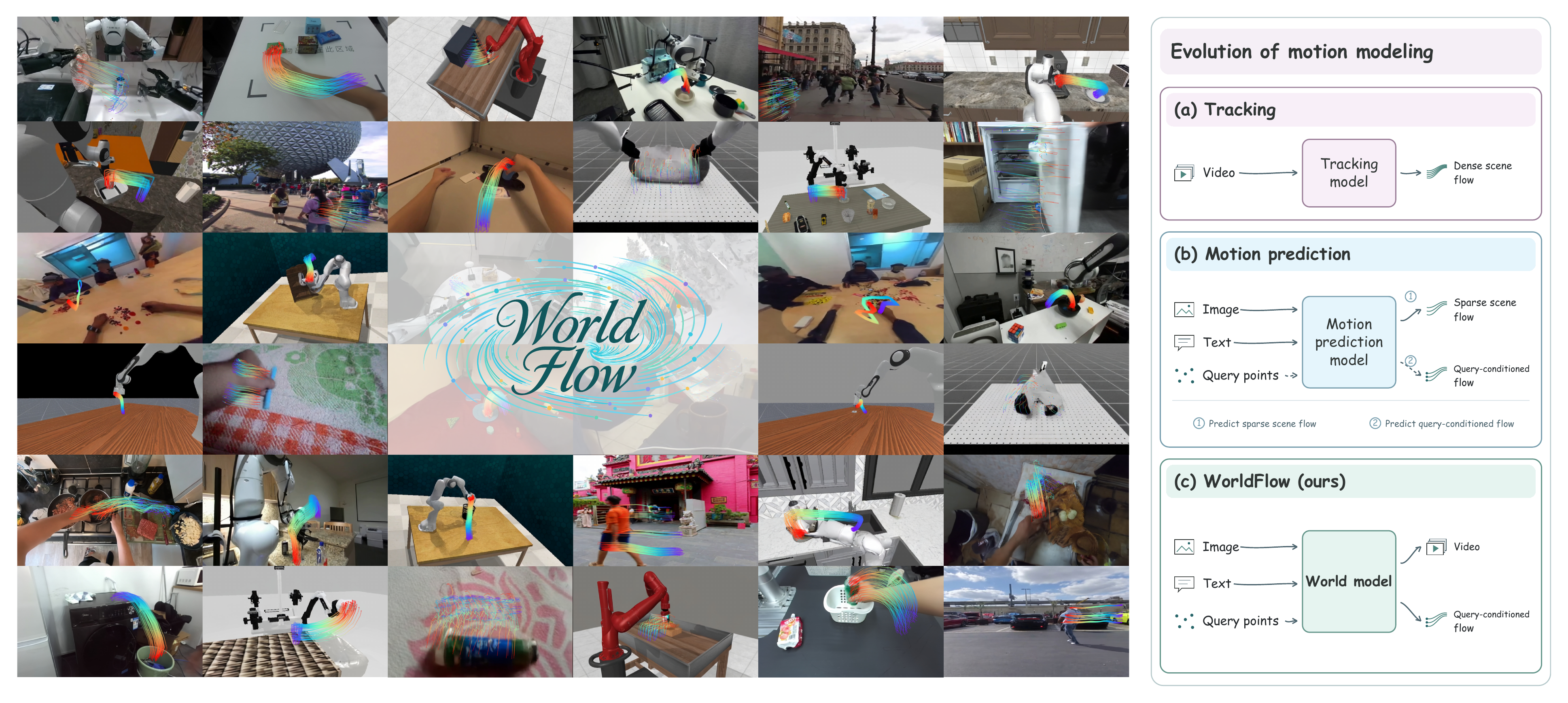}
\end{center}
\vspace{-0.15cm}
\begingroup
\captionsetup{type=figure,hypcap=false}
\caption{
    \textbf{3D Flow Predictions and motion modeling paradigms.}
    \textbf{Left:} 3D motion flow predicted by WorldFlow across simulated, robotic manipulation, human-object interaction, and in-the-wild environments.
    \textbf{Right:} (a) Tracking models recover dense scene flow from video.
    (b) Motion prediction models forecast future 3D flow from an image and text.
    Existing formulations can be broadly divided into sparse scene flow prediction and query-conditioned flow prediction.
    (c) \method{} is a world model conditioned on an image, text, and query points that jointly generates future video and query-conditioned 3D flow.
}
\label{fig:teaser}
\endgroup

\section{Introduction}
\label{sec:introduction}

The philosophy of Heraclitus is often distilled into the
aphorism \textit{panta rhei}---everything flows.
In the physical world, appearances and spatial configurations
continually change, while objects maintain continuity over time.
An embodied world model should capture both aspects.
Recent generative video models~\citep{wan2025wan,seedance2026seedance2}
predict increasingly coherent visual futures, but express
these predictions in pixels, leaving the underlying 3D
dynamics implicit.
World action models~\citep{zhou2026zerowam,yuan2026fastwam,nvidia2026cosmos3,ye2026dreamzero}
connect future prediction to robotic behavior by incorporating
actions into world modeling.
However, action supervision is costly to collect at scale,
and action representations are often tied to particular
robot embodiments, complicating cross-platform transfer.
These limitations motivate extending world modeling into
the space of explicit 3D motion flow: a geometric representation
of physical change that can connect visual prediction to
planning and control across embodiments.

A complementary line of research approaches this space
through motion modeling, progressing from tracking motion
in observed videos~\citep{lu2026track4world,luo20264rc} to predicting future trajectories
from current observations~\citep{zhang2026molmomotion,lee2026mu_0}.
For generative motion prediction, producing dense 3D
scene flow is computationally expensive.
Existing scene-level methods instead predict motion
on coarse spatial grids~\citep{lee2026tracegen,li2026egowam}.
However, sparse scene flow still allocates motion tokens
to task-irrelevant regions, while its reduced spatial
resolution limits the fidelity of fine motion boundaries.
Query-conditioned prediction offers a more targeted
alternative, allocating motion tokens to specified 3D
locations in task-critical regions while retaining global
scene context as conditioning information.
Yet learning generalizable dynamics at scale faces two
challenges: high-quality 3D trajectory annotations are costly
to acquire, while flow-only supervision provides a limited
learning signal for physical dynamics, potentially limiting
generalization across scenes and interactions.
Jointly modeling visual evolution and explicit motion
offers a natural way to enrich this supervision,
bringing motion prediction into a generative world model.

To this end, we introduce \method{}, a generative world model
that jointly predicts future video and query-conditioned
3D object trajectories through a unified
Mixture-of-Transformers (MoT) backbone.
Given an observed image, a language instruction, and sparse
3D query points, \method{} forecasts motion at the queried
locations while using the full scene as visual context.
We represent trajectories as continuous tokens encoding
normalized frame-to-frame 3D displacements, enriched with
initial point geometry and spatiotemporally aligned with
video tokens.
A unified rectified-flow objective jointly denoises both
modalities, coupling trajectory learning with dense visual
supervision of the same future.

To support training at scale, we assemble a heterogeneous
corpus spanning simulated and real-world domains, comprising
approximately 1.44 million object-level samples---equivalent
to 1,224 hours of standardized 49-frame sequences at 16 FPS.
We combine existing motion annotations with automated
trajectory extraction, unify their representations, and
apply source-specific quality control and fixed domain-level
normalization for mixed-data training.
We further introduce WorldFlowBench, comprising three
evaluation sets, including SpatialCog for spatial reasoning
in future motion prediction.
On WorldFlowBench, \method{} achieves the lowest ADE and FDE
across all three evaluation sets among non-static methods.
Compared with its flow-only variant, joint video-flow
generation reduces ADE by 48.5\% on Sim7 and 55.9\% on
SpatialCog, with a 68.5\% reduction in FDE on SpatialCog.
These results support the value of complementary visual
supervision for accurate 3D motion forecasting.

Our contributions are:
\begin{itemize}
    \item We introduce \method{}, a generative world model
    that jointly predicts video and query-conditioned 3D
    object motion through a unified MoT backbone, combining
    explicit trajectory supervision with dense visual
    supervision.

    \item We construct a heterogeneous corpus of approximately
    1.44 million object-level samples from simulated and
    real-world data, using unified motion representations
    and source-specific quality control to support
    large-scale training.

    \item We introduce WorldFlowBench, comprising three
    evaluation sets including SpatialCog for spatial
    reasoning, and demonstrate improvements over existing
    motion prediction methods and
    video-generation-then-tracking pipelines.
\end{itemize}
\section{Related Work}
\label{sec:related}

\subsection{Predictive World Models: From Pixels to Actions}
A fundamental objective of visual world modeling is to predict future physical states from current observations. In the visual domain, large-scale video generators establish rich appearance and motion priors to enable coherent future synthesis~\citep{wan2025wan,seedance2026seedance2}, driven by architectural and training efficiencies~\citep{huang2025selfforcing,chen2025sanavideo}, and extending toward physical-world prediction, interactive simulation, and semantic control~\citep{nvidia2025cosmos,robbyant2026lingbotworld,chen2026h3world,bian2025videoasprompt}. To bridge visual futures with robotic behavior, World Action Models (WAMs) directly incorporate the robot's action space into predictive modeling, jointly forecasting future observations and actions to support end-to-end policy learning and interactive closed-loop control~\citep{ye2026dreamzero,li2026lingbotva}, while expanding toward in-context task generalization~\citep{zhou2026zerowam}, fast inference~\citep{yuan2026fastwam}, and cross-embodiment transfer~\citep{li2026egowam}. Paralleling these efforts, unified multimodal models integrate visual understanding with multi-modal generation~\citep{deng2025bagel}, paving the way for comprehensive frameworks that jointly process language, pixels, and actions for physical-world modeling~\citep{nvidia2026cosmos3}.

\subsection{3D Motion Modeling: From Tracking to Forecasting}
Point tracking stands as a classic paradigm focused on resolving dense, pixel-level correspondences and spatial linkages over observable video sequences~\citep{lu2026track4world,karhade2026any4d,karaev23cotracker,Karaev_2025_ICCV,Xiao_2025_ICCV,tapip3d,luo20264rc}. Although these trackers provide direct measurements of surface history, they operate strictly on immediate visual evidence and remain susceptible to severe error accumulation under complex occlusions or non-rigid deformations. Concurrently, predictive motion forecasting has recently emerged to offer fresh research motivations tailored for downstream embodied and robotic applications. As a generative geometric interface, 3D motion flow bypasses immediate frame matching to predict future physical dynamics directly across objects and embodiments.
Early work, General Flow~\citep{yuan2024general}, learned
language-conditioned 3D point trajectories from human RGB-D videos
as transferable manipulation affordances.
NovaFlow~\citep{li2025novaflow} and
Dream2Flow~\citep{dharmarajan2025dream2flow} leveraged video generation
to obtain motion priors, recovering actionable 3D object flow
through tracking and geometric reconstruction.
Alongside these video-based pipelines, direct motion models
predict future geometry without synthesizing videos.
TraceGen~\citep{lee2026tracegen} models compact 3D traces,
while PointWorld~\citep{huang2026pointworld} predicts
action-conditioned full-scene flow and
ObjectForesight~\citep{soraki2026objectforesight} forecasts
rigid-object pose trajectories.
More recently, RoboFlow4D~\citep{lin2026roboflow4d},
$\mu_0$~\citep{lee2026mu_0}, and
MolmoMotion~\citep{zhang2026molmomotion} advanced
flow-guided planning, interaction-trace generation,
and language-conditioned object-point forecasting, respectively.
Flow representations have also become integral to policy
learning and visual world modeling:
LaMP~\citep{wang2026lamp} introduces a latent 3D motion prior
for action generation,
EgoWAM~\citep{li2026egowam} studies 3D flow as a world-prediction
target for human-to-robot transfer, and
Hydra-0~\citep{li2026hydra} uses image-plane action flow
to condition video prediction and support robot control.
\section{WorldFlow}
\label{sec:method}

\subsection{Problem Formulation}
\label{sec:problem}

Given a language instruction $c$, an observed RGB frame $I_0$,
and 3D query points
$\mathcal{Q}=\{\mathbf{p}_{0,i}\}_{i=1}^{N}$
with corresponding image coordinates
$\mathcal{S}_0=\{\mathbf{s}_{0,i}\}_{i=1}^{N}$,
\method{} jointly predicts future RGB frames $I_{1:T}$
and 3D flow $f_{1:T}$.
All 3D positions are expressed in the initial camera
coordinate system.
We define flow as the frame-to-frame displacement
of each query point:
\begin{equation}
f_{t,i}
=
\mathbf{p}_{t,i}-\mathbf{p}_{t-1,i},
\qquad t=1,\ldots,T.
\label{eq:flow-definition}
\end{equation}
The corresponding trajectories are recovered by accumulation:
\begin{equation}
\mathbf{p}_{t,i}
=
\mathbf{p}_{0,i}+\sum_{\tau=1}^{t}f_{\tau,i}.
\label{eq:trajectory-reconstruction}
\end{equation}
Our objective is to model the conditional joint distribution
\begin{equation}
p_\theta\left(
I_{1:T},f_{1:T}
\mid I_0,\mathcal{Q},\mathcal{S}_0,c
\right).
\label{eq:joint-distribution}
\end{equation}

\subsection{Architecture}
\label{sec:architecture}

\begin{figure}[t]
    \centering
    \includegraphics[width=\linewidth]{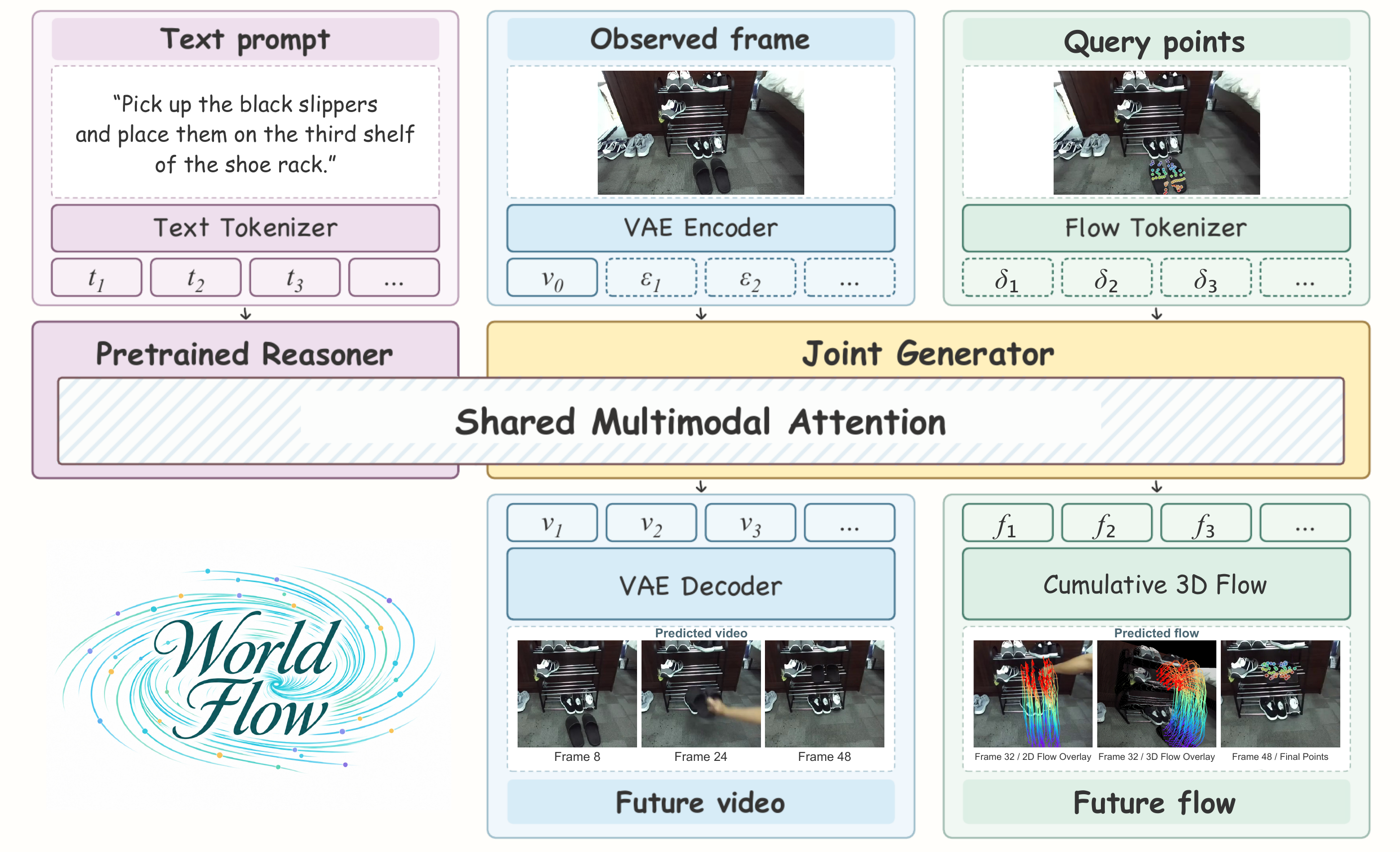}
    \caption{
        Overview of \method{}.
        Conditioned on text tokens $t_j$ through the pretrained
        reasoner, clean observed-frame latents $v_0$, and query-point geometry, the generator jointly denoises video and flow from noise $\epsilon_j$ and $\delta_k$ through shared multimodal attention.
        Generated video latents $v_j$ are decoded by the VAE,
        while generated 3D flow $f_t$ is rescaled and
        accumulated from the initial query positions
        to recover metric trajectories.
        The 2D flow overlay shows generated 3D trajectories projected onto generated video frames.
        The 3D flow overlay shows generated 3D trajectories overlaid on a scene point cloud reconstructed from
        the initial frame's ground-truth depth.
    }
    \label{fig:overview}
\end{figure}

Our central design introduces 3D flow tokens as an additional
generative modality for joint video-motion prediction.
Built on the Cosmos-Nano framework~\citep{nvidia2026cosmos3}, \method{} retains
the pretrained reasoner and extends the generator to
jointly synthesize future video and 3D flow
(Figure~\ref{fig:overview}).
The reasoner processes the language instruction, while
the video VAE encodes the observed frame into clean latents.
The flow tokenizer embeds noisy displacements together
with initial query-point geometry and aligns their
spatiotemporal positional coordinates with the video
token grid (Section~\ref{sec:tokenizer}).

Within the Mixture-of-Transformers (MoT) architecture,
video and flow tokens interact through full attention
and attend to the reasoner's causal context.
At inference, future video latents and flow are initialized
from noise and jointly denoised, conditioned on the clean
observed-frame latents, language context, and query geometry.
The VAE decodes the generated video latents into RGB frames,
while the predicted flow is rescaled and accumulated
to recover 3D trajectories.
This joint generation process produces explicit motion
alongside video without a separate tracking stage.

\subsection{3D Flow Tokenization}
\label{sec:tokenizer}

We represent each query point's displacement at each future
frame as a continuous-valued token, augmented with initial
3D geometry and spatiotemporal positional encoding.

\paragraph{Normalization and sequence construction.}
To balance motion scales across domains, we normalize
displacements by a domain-specific scalar estimated
from training data and fixed throughout training:
\begin{equation}
\bar{f}_{t,i}=\frac{f_{t,i}}{a_d},
\qquad a_d>0.
\label{eq:motion-normalization}
\end{equation}
The normalized displacements are arranged in frame-major
order, yielding $TN$ three-dimensional vectors for
token construction.

\paragraph{Query point geometry encoder.}
We encode each query point's absolute initial position
and its position relative to the query-set centroid
using two parallel encoders.
Each applies coordinate normalization, fixed Fourier
features, and an MLP.
Their outputs are concatenated and fused:
\begin{equation}
\mathbf{e}^{\mathrm{3D}}_i
=
\mathrm{MLP}
\left(
\left[
E_{\mathrm{abs}}(\mathbf{p}_{0,i});
E_{\mathrm{rel}}(\mathbf{p}_{0,i}-\boldsymbol{\mu})
\right]
\right),
\label{eq:point-pe}
\end{equation}
where $E_{\mathrm{abs}}$ and $E_{\mathrm{rel}}$ denote
the two encoders, and $\boldsymbol{\mu}$ is the query-set
centroid.
The geometry embedding is shared across all future
tokens of the same query point.

\paragraph{Flow token construction.}
A linear projection maps noisy normalized displacements
into the generator's hidden space, where they are combined
with geometry and conditioning embeddings:
\begin{equation}
\mathbf{h}_{t,i}^{(\sigma)}
=
\mathrm{Linear}(\bar{f}_{t,i}^{(\sigma)})
+
\mathbf{e}^{\mathrm{3D}}_i
+
\mathbf{e}_{\mathrm{cond}}(\sigma).
\label{eq:flow-embedding}
\end{equation}
Here, $\sigma$ denotes the denoising noise level,
distinct from the physical frame index $t$.
The conditioning embedding combines a learned
flow-modality embedding with the noise-level embedding.

\paragraph{Spatiotemporal alignment.}
We assign flow tokens rotary positional coordinates
using the same temporal and spatial conventions as
video tokens.
Each flow token is anchored to its future frame
and the query point's initial image patch:
\begin{equation}
\mathbf{r}_{t,i}
=
\left(
\frac{t}{q_t},
\left\lfloor\frac{v_{0,i}}{q_s p_s}\right\rfloor,
\left\lfloor\frac{u_{0,i}}{q_s p_s}\right\rfloor
\right),
\label{eq:flow-rope-coordinates}
\end{equation}
where $(u_{0,i},v_{0,i})=\mathbf{s}_{0,i}$ denotes
the initial image coordinates, $q_t$ and $q_s$ are
the temporal and spatial VAE compression factors,
and $p_s$ is the latent patch size.
These coordinates determine the rotary transformations
of attention queries and keys:
\begin{equation}
\widetilde{\mathbf{q}}_{t,i}
=
R(\mathbf{r}_{t,i})\mathbf{q}_{t,i},
\qquad
\widetilde{\mathbf{k}}_{t,i}
=
R(\mathbf{r}_{t,i})\mathbf{k}_{t,i}.
\label{eq:flow-rope}
\end{equation}
The shared positional convention encodes relative
spatiotemporal relationships between flow and video
tokens within multimodal attention.

\subsection{Training Objectives}
\label{sec:training}

\subsubsection{Joint flow matching}
We jointly train video and 3D flow generation using
a weighted rectified-flow objective:
\begin{equation}
\mathcal{L}_{\mathrm{FM}}
=
\lambda_{\mathrm{video}}\mathcal{L}_{\mathrm{video}}
+
\lambda_{\mathrm{flow}}\mathcal{L}_{\mathrm{flow}},
\label{eq:joint-objective}
\end{equation}
where $\mathcal{L}_{\mathrm{video}}$ and
$\mathcal{L}_{\mathrm{flow}}$ supervise vector-field
prediction for video latents and normalized 3D flow,
respectively.
Both modalities share a per-sample noise level and
are jointly denoised.
Losses are computed separately for each modality,
excluding conditioning positions from supervision.

\subsubsection{Auxiliary motion objectives}
To complement vector-field supervision with explicit
constraints on predicted motion, we apply auxiliary
objectives to clean normalized flow reconstructed
from the predicted vector field.
These objectives penalize frame-to-frame displacement
error, cumulative displacement direction error,
and relative endpoint error, encouraging both local
accuracy and trajectory-level consistency.

Direction errors are weighted by ground-truth displacement
magnitude, while points with small net displacement
are excluded from endpoint supervision.
All auxiliary terms operate in normalized coordinates
and use noise-dependent weighting to emphasize
lower-noise samples.
The complete objective is
\begin{equation}
\mathcal{L}
=
\mathcal{L}_{\mathrm{FM}}
+
\lambda_{\mathrm{dir}}\mathcal{L}_{\mathrm{dir}}
+
\lambda_{\mathrm{vmag}}\mathcal{L}_{\mathrm{vmag}}
+
\lambda_{\mathrm{end}}\mathcal{L}_{\mathrm{end}},
\label{eq:complete-objective}
\end{equation}
where $\mathcal{L}_{\mathrm{dir}}$,
$\mathcal{L}_{\mathrm{vmag}}$, and
$\mathcal{L}_{\mathrm{end}}$ denote the direction,
displacement, and endpoint losses, respectively.
Detailed formulations and derivations are provided
in the supplementary material, and their contribution
is evaluated through ablation experiments.
\section{Experiments}
\label{sec:experiments}

\subsection{Datasets}

We assemble a heterogeneous corpus of paired videos
and 3D point trajectories spanning simulated and real-world
interactions.
The training corpus contains approximately 1.44 million
object-level samples, equivalent to 1,224 hours of
standardized 49-frame sequences at 16 FPS.
The data are organized into four collections:
\begin{itemize}
    \raggedright
    \item \textbf{Sim7} combines seven simulation datasets:
    ManiSkill~\citep{tao2024maniskill3},
    MIKASA~\citep{cherepanov2026memory},
    MetaWorld~\citep{yu2020meta},
    MetaFold~\citep{chen2025metafold},
    RLBench~\citep{james2020rlbench},
    RoboCasa~\citep{nasiriany2024robocasa}, and
    RoboTwin~\citep{mu2025robotwin},
    covering diverse robotic interactions and manipulation tasks.

    \item \textbf{Real3} comprises
    AgiBot World~\citep{bu2025agibot},
    EPIC-KITCHENS~\citep{damen2018epickitchens}, and
    Something-Something V2~\citep{goyal2017something},
    covering real-world robot and human interactions.

    \item \textbf{SpatialCog} is our self-collected dataset
    of tabletop and household interactions, with an emphasis
    on spatial relations in instruction-conditioned motion.

    \item \textbf{MolmoMotion-1M}~\citep{zhang2026molmomotion}
    contributes curated 3D motion annotations from six sources:
    EgoDex~\citep{hoque2025egodex},
    Xperience~\citep{ropedia2026xperience},
    YT-VIS~\citep{Yang_2019_ICCV},
    Stereo4D~\citep{Jin_2025_CVPR},
    DROID~\citep{khazatsky2024droid}, and
    MolmoSpaces~\citep{kim2026molmospaces}.
\end{itemize}

\paragraph{Annotation and standardization.}
For Sim7 and Real3, we generate captions with
Gemini-2.5 Pro~\citep{comanici2025gemini}, obtain object
masks with SAM 2~\citep{ravi2024sam2}, and extract 3D
point trajectories using Track4World~\citep{lu2026track4world}.
SpatialCog pairs manually annotated captions with
Track4World trajectories.
For MolmoMotion-1M, we reuse the released motion
annotations, apply source-specific geometry, scale,
and trajectory-validity checks, and standardize the
retained sequences into 49-frame clips.
All collections use a shared representation consisting
of video, a language instruction, initial 3D query points,
and frame-to-frame 3D displacements.
Additional processing details and per-source statistics
are provided in the supplementary material.

\paragraph{Evaluation split.}
We introduce WorldFlowBench, a benchmark comprising
1,678 held-out evaluation samples: 396 from Sim7,
882 from Real3, and 400 from SpatialCog.
The three subsets cover simulated interactions,
real-world motion, and spatial-relation reasoning,
respectively.
All evaluation samples are excluded from training.

\subsection{Training Configuration and Strategy}

\paragraph{Implementation details.}
We train on 49-frame clips at $288\times480$ resolution
with $N=64$ query points.
The video VAE uses temporal and spatial compression
factors of $q_t=4$ and $q_s=16$, followed by
$2\times2$ latent patchification ($p_s=2$).
Each clip therefore contains a $13\times9\times15$
video-token grid and $48\times64$ flow tokens.
Training uses 64 NVIDIA H800 GPUs across eight nodes,
with a per-GPU batch size of two and no gradient
accumulation, yielding a global batch size of 128.

\paragraph{Training strategy.}
We adopt a two-stage training schedule.
The first stage initializes the model from pretrained
Cosmos-Nano checkpoints and trains on the SpatialCog
training split for 32,000 steps.
The learning rate follows linear warmup and cosine decay,
with a peak of $10^{-4}$.
The resulting exponential moving average (EMA) weights
initialize the second stage.

The second stage trains on the heterogeneous data mixture
for 51,100 steps across three curriculum phases:
simulation-focused training, mixed real-world adaptation,
and refinement on the SpatialCog training split.
During real-world adaptation, sampling gradually shifts
from MolmoMotion to Real3.
Each phase prioritizes its target domain while retaining
cross-domain replay, with smooth transitions between
sampling distributions.
After warmup, the learning rate reaches $10^{-4}$,
transitions to an $8\times10^{-5}$ plateau during
real-world adaptation, and decays through late adaptation
and final refinement.

We expand task prompts into structured instructions
specifying data origin, camera motion, actor type,
and environment.
For classifier-free guidance, text conditioning is
dropped with probability $0.1$, while visual and
geometric conditioning is retained.
At inference, we use 40 rectified-flow sampling steps
with a guidance scale of $2$.

\subsection{Flow Benchmark}

\subsubsection{Evaluation Metrics}
We evaluate 3D point trajectories in metric coordinates.
Average displacement error (ADE) measures the mean
Euclidean position error across query points and future
frames, while final displacement error (FDE) measures
the mean position error at the final frame.
Both are reported in meters.
PWT measures the fraction of point--frame pairs whose
position error falls within a given distance threshold.
We report its average over thresholds of 1, 2, and 5~cm
to assess trajectory accuracy at fine spatial scales.

\subsubsection{Quantitative Evaluation on WorldFlowBench}
\label{sec:worldflowbench_eval}

We evaluate all learning-based methods under a single-frame
protocol, where one RGB observation and a language instruction
are provided to forecast a common future horizon without motion
history. All predictions are converted into metric 3D point
trajectories in the initial camera coordinate system. We
additionally report a Static reference that predicts zero flow
and therefore keeps every query point at its initial position
throughout the prediction horizon.

We compare against two families of learned baselines. The
video-generation-then-tracking family includes
NovaFlow~\citep{li2025novaflow}, as well as pipelines built on
Cosmos3-Nano~\citep{nvidia2026cosmos3} or
LingBot-Video-MoE~\citep{robbyant2026lingbotworld}, paired with
Track4World (T4W)~\citep{lu2026track4world} or
4RC~\citep{luo20264rc} for 3D tracking. The direct flow-generation
family includes MolmoMotion-AR-H1~\citep{zhang2026molmomotion},
$\mu_0$~\citep{lee2026mu_0}, and
EgoWAM~\citep{li2026egowam}.

As shown in Table~\ref{tab:worldflowbench}, \method{} achieves
the lowest ADE and FDE among all evaluated learning-based methods
on all three subsets. Relative to the strongest external learned
baseline on each subset, it reduces ADE by approximately
46.9\% on Sim7, 4.7\% on Real3, and 54.2\% on SpatialCog.
It also achieves the highest PWT among learning-based methods
on Sim7 and SpatialCog. On Real3, \method{} obtains a PWT of
0.311, compared with the best value of 0.332, while still
delivering the lowest ADE and FDE. These results demonstrate
consistent improvements in trajectory accuracy across simulated
interactions, real-world motion, and spatial-reasoning tasks.

\begin{figure}[t]
    \centering
    \includegraphics[width=\textwidth]{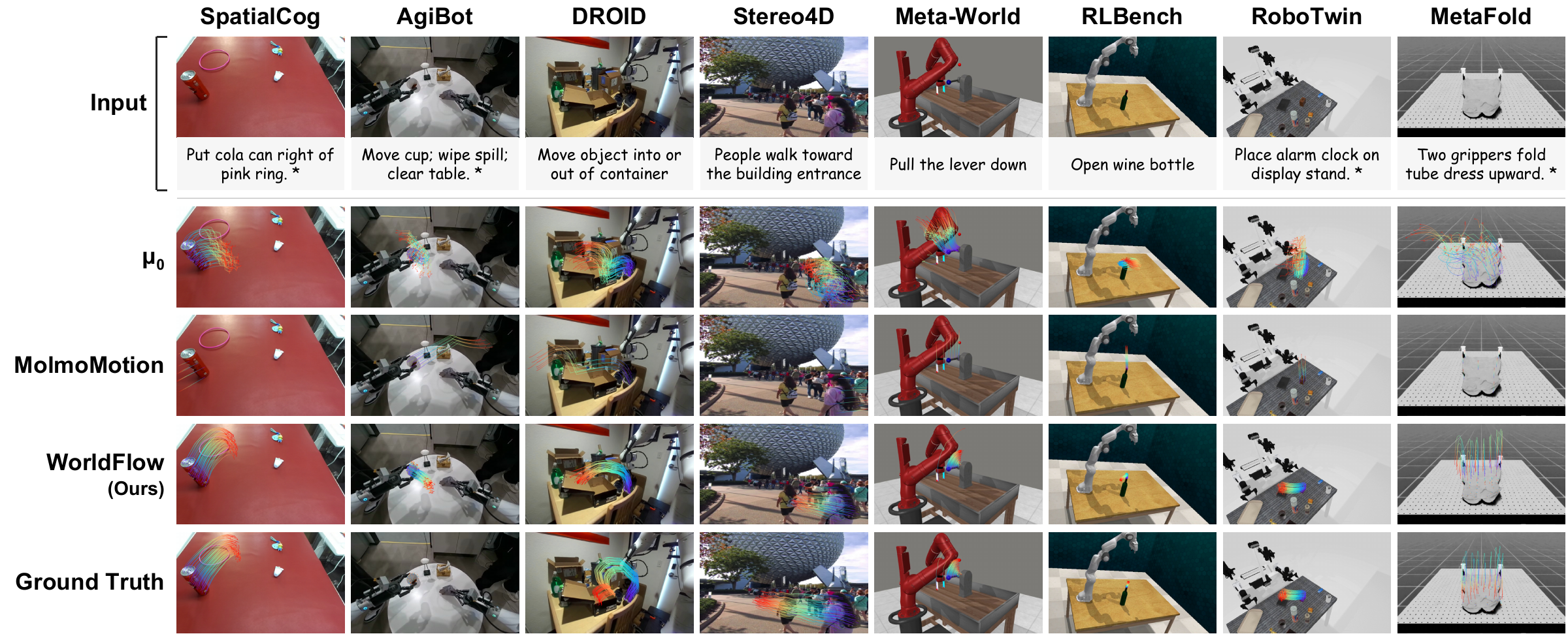}
    \caption{
        \textbf{Qualitative comparison on diverse real and simulated scenes.}
        We visualize predicted 3D query-point trajectories from $\mu_0$, MolmoMotion, and \method{} across SpatialCog, AgiBot, DROID, Stereo4D, Meta-World, RLBench, RoboTwin, and MetaFold.
        The trajectories are projected onto the corresponding input frames for visualization.
    }
    \label{fig:qualitative-comparison}
\end{figure}

\begin{table*}[t]
\centering
\small
\setlength{\tabcolsep}{4pt}
\renewcommand{\arraystretch}{1.12}
\resizebox{\textwidth}{!}{%
\begin{tabular}{@{}ll ccc ccc ccc@{}}
\toprule
\multirow{2}{*}{\textbf{Paradigm}} &
\multirow{2}{*}{\textbf{Model}} &
\multicolumn{3}{c}{\textbf{Sim7}} &
\multicolumn{3}{c}{\textbf{Real3}} &
\multicolumn{3}{c}{\textbf{SpatialCog}} \\
\cmidrule(lr){3-5}
\cmidrule(lr){6-8}
\cmidrule(lr){9-11}
& &
ADE$\downarrow$ & FDE$\downarrow$ & PWT$\uparrow$ &
ADE$\downarrow$ & FDE$\downarrow$ & PWT$\uparrow$ &
ADE$\downarrow$ & FDE$\downarrow$ & PWT$\uparrow$ \\
\midrule
Non-parametric
& Static
& 0.096 & 0.201 & 0.573
& 0.062 & 0.086 & 0.464
& 0.143 & 0.278 & 0.412 \\
\midrule
\multirow{5}{*}{\shortstack[l]{Video generation\\+ 3D tracking}}
& NovaFlow
& 0.196 & 0.318 & 0.289
& 0.095 & 0.129 & 0.293
& 0.165 & 0.295 & 0.320 \\
& Cosmos3-Nano + T4W
& 0.101 & 0.202 & 0.451
& 0.088 & 0.133 & 0.310
& 0.141 & 0.257 & 0.398 \\
& Cosmos3-Nano + 4RC
& 0.096 & 0.196 & \underline{0.531}
& 0.085 & 0.122 & \textbf{0.332}
& 0.141 & 0.263 & 0.401 \\
& LingBot-Video-MoE + T4W
& 0.105 & 0.207 & 0.425
& 0.097 & 0.145 & 0.267
& 0.134 & 0.239 & 0.364 \\
& LingBot-Video-MoE + 4RC
& 0.104 & 0.207 & 0.492
& 0.092 & 0.134 & 0.295
& 0.131 & 0.232 & 0.376 \\
\midrule
\multirow{3}{*}{\shortstack[l]{Direct flow\\generation}}
& MolmoMotion-AR-H1
& 0.176 & 0.323 & 0.280
& 0.185 & 0.335 & 0.222
& 0.282 & 0.533 & 0.173 \\
& $\mu_0$
& 0.186 & 0.307 & 0.069
& 0.116 & 0.156 & 0.148
& 0.213 & 0.270 & 0.055 \\
& EgoWAM
& 0.198 & 0.336 & 0.020
& 0.130 & 0.190 & 0.056
& 0.249 & 0.383 & 0.080 \\
\midrule
\multirow{4}{*}{Ablations}
& \method{} (flow-only)
& 0.099 & 0.206 & 0.495
& 0.101 & 0.152 & \underline{0.320}
& 0.136 & 0.248 & 0.372 \\
& \method{} + T4W
& 0.088 & 0.174 & 0.424
& 0.100 & 0.154 & 0.263
& 0.083 & 0.138 & 0.403 \\
& \method{} + 4RC
& 0.088 & 0.170 & 0.483
& 0.092 & 0.132 & 0.291
& 0.098 & 0.175 & 0.411 \\
& \method{} w/o auxiliary loss
& \underline{0.063} & \underline{0.116} & 0.527
& \underline{0.083} & \underline{0.115} & 0.299
& \underline{0.062} & \underline{0.084} & \underline{0.436} \\
\midrule
\textbf{Ours}
& \method{}
& \textbf{0.051} & \textbf{0.094} & \textbf{0.564}
& \textbf{0.081} & \textbf{0.113} & 0.311
& \textbf{0.060} & \textbf{0.078} & \textbf{0.449} \\
\bottomrule
\end{tabular}%
}
\caption{
\textbf{3D motion forecasting on WorldFlowBench.}
ADE and FDE measure average and final position errors
in meters, respectively; lower is better.
PWT is the mean point--frame hit rate over distance
thresholds of 1, 2, and 5~cm; higher is better.
Bold and underline mark the best and second-best results
in each column among non-static methods, including ablations.
Static is reported as a reference and excluded from ranking.
T4W denotes Track4World.
For \method{} + T4W/4RC, trajectories are extracted from
the videos generated by \method{} using the corresponding
tracker.
}
\label{tab:worldflowbench}
\end{table*}

\subsubsection{Ablation Studies}

\paragraph{Flow-only versus joint generation.}
To evaluate the benefit of jointly modeling 3D motion
and visual evolution, we compare \method{} with a
flow-only variant using the same backbone, training data,
and optimization recipe.
The variant predicts only 3D flow, whereas the full model
jointly generates video and flow.
The full model reduces ADE by 48.5\%, 19.8\%, and 55.9\%
on Sim7, Real3, and SpatialCog, respectively, with lower
FDE on all three subsets.
On SpatialCog, FDE decreases from $0.248$ to
$0.078\,\mathrm{m}$, a 68.5\% reduction.
PWT also improves on Sim7 and SpatialCog, with a small
decrease on Real3.
These results support the value of complementary visual
supervision for learning accurate 3D motion.

\paragraph{Direct flow generation versus post-hoc tracking.}
To evaluate flow as an explicit generative modality,
we compare directly generated trajectories with those
recovered by Track4World and 4RC from videos produced
by the same \method{} checkpoint.
This comparison holds the video generator and conditioning
inputs fixed, contrasting explicit motion generation
with motion recovery from generated pixels.
Direct flow generation achieves lower ADE and FDE
and higher PWT than both tracking pipelines on all
three subsets.
On SpatialCog, it achieves an ADE of $0.060\,\mathrm{m}$,
compared with $0.083\,\mathrm{m}$ for Track4World and
$0.098\,\mathrm{m}$ for 4RC.
These results support incorporating explicit 3D flow
into the world model's generative output rather than
recovering motion through a separate tracking stage.

\paragraph{Auxiliary motion supervision.}
We compare the full model with a variant trained without
auxiliary motion objectives.
These objectives supervise displacement, cumulative
motion direction, and trajectory endpoints through
the reconstructed clean flow.
Adding them improves ADE, FDE, and PWT on all three
subsets, with the largest ADE reduction on Sim7,
from $0.063$ to $0.051\,\mathrm{m}$ (19.0\%).
PWT increases from 0.527 to 0.564 on Sim7,
from 0.299 to 0.311 on Real3, and from 0.436 to 0.449
on SpatialCog.
These consistent improvements indicate that explicit
supervision of clean motion complements the primary
flow-matching objective.




\subsection{Flow-Guided Robotic Manipulation}
\label{sec:flow_guided_manipulation}

We evaluate the capability of 3D motion prediction models in robotic scenarios and their utility in guiding downstream execution through both simulation and real-robot experiments. Within modular robotic manipulation pipelines, we compare \method{} with $\mu_0$~\citep{lee2026mu_0}, which is selected for its competitive offline trajectory prediction performance among reproducible direct motion predictors. Furthermore, RoboLab provides an out-of-distribution (OOD) testbed for \method{} to validate its generalization under novel conditions.

\subsubsection{Simulation on RoboLab}
\label{sec:sim_robot}

We evaluate \method{} on RoboLab~\citep{yang2026robolab},
a public manipulation benchmark providing OOD inference
scenarios for the motion predictor.
We use the pretrained checkpoint without RoboLab-specific
post-training or fine-tuning. Specifically, we adopt the \texttt{Simple} task suite of RoboLab and evaluate all 64 tasks within it under \texttt{Vague}, \texttt{Default}, and \texttt{Specific} instructions. Each task-instruction configuration is tested across three trials, yielding 192 trials per instruction variant and 576 trials per motion predictor under each execution setting.

RoboLab uses a simulated Franka arm with a Robotiq 2F-85
two-finger gripper. Our pipeline comprises grasp acquisition,
flow-guided transport, and placement.
Grasping uses a rule-based policy with geometric checks,
refined offline through GPT-6.1 Sol-assisted iteration
and simulation feedback~\citep{OpenAI2026GPT61Sol},
then fixed for evaluation without online GPT calls.
Qwen3.8-27B~\citep{qwen38} segments the instructed object;
its mask and depth yield a point cloud for sampling
3D query points. Given these points, the image, and the
instruction, the motion predictor generates object
trajectories, which an optimizer converts into end-effector
poses under grasp and kinematic constraints.
The robot tracks these poses through inverse kinematics
and releases the object at the predicted endpoint.
Both predictors are evaluated under two settings using
RoboLab's automated task completion criterion.
Within each setting, perception, grasping, trajectory
optimization, and control modules are shared;
only the motion predictor changes.
\textbf{Orch} executes the full pipeline to measure
end-to-end manipulation success.
\textbf{Attach} replaces physical grasping with
simulator-assisted attachment to prevent grasp failures
and slippage during transport, allowing flow-guided
task completion to be evaluated with reduced
grasp-related interference.

\paragraph{Simulation results.}
Table~\ref{tab:robolab_manipulation} evaluates OOD
manipulation from two perspectives.
Within our pipelines, \method{} outperforms $\mu_0$
across all three instruction variants and both settings.
Under Orch, overall task success increases from
3.82\% (22/576) to 38.54\% (222/576), a gain of
34.72 percentage points.
Under Attach, success increases from 9.20\% (53/576)
to 47.22\% (272/576), a gain of 38.02 percentage points.
These gains demonstrate more effective transfer of
predicted object motion to OOD manipulation than $\mu_0$.
For system-level context, WorldFlow-Orch also exceeds
the published success rates of the selected VLA and
WAM policies across all three instruction variants.
Its overall success rate is 3.23 percentage points
higher than Cosmos3-Nano-Policy, the strongest of
these references.
Together, these results support the practical value
of WorldFlow's motion predictions for robotic
manipulation in OOD environments.

\begin{table*}[t]
\centering
\small
\setlength{\tabcolsep}{5pt}
\renewcommand{\arraystretch}{1.05}
\caption{
\textbf{Task success on RoboLab Simple.}
Published policy results from the RoboLab leaderboard are
reported under their original configurations for reference.
Motion predictors are compared under two settings:
Orch (full manipulation) and Attach (simulator-assisted
object retention).
Bold and underline mark the highest and second-highest
values per column across the displayed settings.
}
\label{tab:robolab_manipulation}
\begin{tabular*}{\textwidth}{@{\extracolsep{\fill}}lcccc@{}}
\toprule
Method & Vague & Default & Specific & Overall \\
\midrule
Cosmos3-Nano-Policy~\citep{nvidia2026cosmos3}
& 23.28\% & 40.63\% & 42.03\% & 35.31\% \\
$\pi_{0.5}$~\citep{black2025pi05}
& 16.25\% & 29.69\% & 29.84\% & 25.26\% \\
DreamZero~\citep{ye2026dreamzero}
& 15.00\% & 26.09\% & 25.63\% & 22.24\% \\
GR00T N1.6~\citep{gr00tn1_2025}
& 7.19\% & 8.75\% & 7.50\% & 7.81\% \\
\midrule
$\mu_0$-Orch
& 3.13\% & 3.65\% & 4.69\% & 3.82\% \\
WorldFlow-Orch
& \underline{28.13\%} & \underline{44.79\%}
& \underline{42.71\%} & \underline{38.54\%} \\
\cmidrule(lr){1-5}
$\mu_0$-Attach
& 8.33\% & 8.33\% & 10.94\% & 9.20\% \\
WorldFlow-Attach
& \textbf{32.29\%} & \textbf{52.60\%}
& \textbf{56.77\%} & \textbf{47.22\%} \\
\bottomrule
\end{tabular*}
\end{table*}

\subsubsection{Real-Robot Manipulation}
\label{sec:real_robot}

\paragraph{Setup and execution.}
Our physical setup uses a 7-DoF xArm, an XHand dexterous hand,
and a ZED Mini RGB-D camera. Given an RGB-D observation and a
language instruction, Qwen3.8-27B~\citep{qwen38} identifies
and segments the target object. We use the resulting mask
and depth image to extract a 3D point cloud and sample query
points for trajectory prediction. T(R,O) Grasp~\citep{fei2025t}
generates the initial dexterous grasp pose. Following
Dream2Flow~\citep{dharmarajan2025dream2flow}, we fit a locally
rigid patch near the grasp to the predicted object trajectories
to obtain the corresponding arm trajectory.
We evaluate 12 tasks covering pick and place, stacking,
and relative placement, with three randomized trials
per task and 36 executions per method.
We evaluate flow prediction and robot execution separately.
Flow prediction success is manually assessed based on
whether the predicted object motion is plausible for
the given instruction. Robot execution succeeds if the
full pipeline completes the task on the physical robot.
In each paired trial, we vary only the flow predictor
and keep the initial layout, instruction, grasp result,
and control settings identical.

\paragraph{Real-robot results.}
As shown in Table~\ref{tab:real_robot}, \method{} achieves
97.2\% flow prediction success (35/36) and 66.7\%
end-to-end execution success (24/36), compared with
25.0\% (9/36) and 11.1\% (4/36) for $\mu_0$, respectively.
Execution failures despite plausible motion predictions
arise from grasping, trajectory optimization, and unstable
placement during stacking.
With the grasping and control modules held fixed,
these results show that \method{} provides more reliable
object-motion references, translating better flow
predictions into higher real-world task completion.

\begin{table*}[t]
\centering
\small
\setlength{\tabcolsep}{6pt}
\renewcommand{\arraystretch}{1.15}
\resizebox{\textwidth}{!}{%
\begin{tabular}{@{}llcccc@{}}
\toprule
\multirow{2}{*}{\textbf{Group}}
& \multirow{2}{*}{\textbf{Task Caption}}
& \multicolumn{2}{c}{\textbf{Flow Prediction}}
& \multicolumn{2}{c}{\textbf{Robot Execution}} \\
\cmidrule(lr){3-4}
\cmidrule(lr){5-6}
& & \method{} & $\mu_0$ & \method{} & $\mu_0$ \\
\midrule

Pick and Place
& Place the watermelon doll into the small pink plate.
& \textbf{3/3} & 0/3 & \textbf{3/3} & 0/3 \\
& Place the oolong tea into the large blue tray.
& \textbf{3/3} & 1/3 & \textbf{2/3} & 0/3 \\
& Place the potato chips into the dusty pink basket.
& \textbf{3/3} & 1/3 & \textbf{2/3} & 0/3 \\
& Place the yellow toy inside the red circle.
& \textbf{3/3} & 1/3 & \textbf{2/3} & 1/3 \\
\midrule

Stacking
& Place the green biscuit box on the translucent box.
& \textbf{3/3} & 1/3 & \textbf{3/3} & 0/3 \\
& Place the can of Red Bull on the translucent box.
& \textbf{3/3} & 1/3 & \textbf{2/3} & 1/3 \\
& Place the beef curry on the paper towel.
& \textbf{3/3} & 1/3 & \textbf{1/3} & 0/3 \\
& Place the yellow toy on the paper towel.
& \textbf{3/3} & 2/3 & 0/3 & 0/3 \\
\midrule

Relative Placement
& Place the sunscreen in front of the tissue.
& \textbf{2/3} & 0/3 & \textbf{1/3} & 0/3 \\
& Place the potato chips behind the Oreo box.
& \textbf{3/3} & 0/3 & \textbf{2/3} & 1/3 \\
& Place the watermelon doll to the left of the cola.
& \textbf{3/3} & 1/3 & \textbf{3/3} & 1/3 \\
& Place the oolong tea to the right of the small blue toy car.
& \textbf{3/3} & 0/3 & \textbf{3/3} & 0/3 \\
\midrule

\multicolumn{2}{@{}l}{\textbf{Overall success rate}}
& \textbf{97.2\%} & 25.0\%
& \textbf{66.7\%} & 11.1\% \\
\bottomrule
\end{tabular}%
}
\caption{
\textbf{Real-world manipulation with different flow predictors.}
Each task is evaluated over three trials, yielding 36 trials
per method. Flow prediction and robot execution are evaluated
separately for each trial. Both methods use matched initial
layouts, identical grasp results, and the same trajectory
optimization and execution settings; only the flow predictor
changes. Bold indicates the higher success rate.
}
\label{tab:real_robot}
\end{table*}
\section{Conclusion}
\label{sec:conclusion}

We presented \method{}, a world model that jointly generates
future video and metric 3D flow from a single observation,
a language instruction, and 3D query points.
Continuous flow tokens are grounded in video coordinates
and initial point geometry, enabling both modalities to be
jointly denoised by a shared rectified-flow generator.
Experiments across diverse embodied domains demonstrate
improved motion forecasting over direct flow predictors
and video-generation-then-tracking pipelines.

The current formulation requires predefined query points
and inherits noise from automatically generated trajectory
supervision.
Future work will explore automatic query selection,
multimodal futures, and joint generation of 3D flow and
actions to study the benefits of explicit motion
representations for policy learning.

\section*{AI Use Statement}
We used generative AI tools to assist with translation,
provide feedback on experimental methodology, and
interpret experimental results.
Gemini 2.5 Pro was used to generate captions for
Sim7 and Real3, as described in the dataset section.
We additionally used AI tools to assist with manuscript
drafting, language editing, organization, and LaTeX
formatting.
The authors reviewed the AI-assisted content and checked
technical descriptions, numerical comparisons, and
conclusions against the methodology and experimental
records.
The authors take full responsibility for the final
content of this work, including all AI-assisted text,
claims, and artifacts.


\bibliographystyle{rbs}
\bibliography{references}

\clearpage
\appendix
\setcounter{figure}{0}
\setcounter{table}{0}
\renewcommand{\thefigure}{\Roman{figure}}
\renewcommand{\thetable}{\Roman{table}}
\renewcommand{\theHfigure}{appendix.\Roman{figure}}
\renewcommand{\theHtable}{appendix.\Roman{table}}

\begin{center}
    {\Large\bf Supplementary Material for \\[2pt]
    ``WorldFlow: Joint Video and 3D Motion Prediction via Large-Scale Pretraining''}
\end{center}
\vspace{1em}


\section{Data Processing Details and Dataset Statistics}
\label{sec:supp-data}

\subsection{Data Processing Details}

The main paper provides an overview of the data pipeline;
here, we describe the source-specific processing details.
For Sim7, Real3, and SpatialCog, we extract trajectories
directly from the original videos using a standardized
configuration of 49 RGB frames and 64 query points per
object record.
Tracked pixels are back-projected using depth and camera
intrinsics, and camera poses transform the resulting 3D
points into the first-frame camera coordinate system.
The object mask in the first frame defines the visible
partial surface from which the query points are selected.
RGB frames and point trajectories are generated at the
same temporal indices and therefore require no subsequent
temporal realignment.

For MolmoMotion-1M, we reuse the released trajectories and
associated geometry rather than extracting trajectories
from the source videos.
We transform the trajectories into the same first-frame
camera coordinate system and standardize their point
counts and temporal lengths.
Records with fewer than eight valid object-point candidates
are discarded.
When more than 64 candidates are available, we use
deterministic voxel-uniform sampling to maintain spatial
coverage; for records with 8--63 candidates, we sample
64 points with replacement using a fixed random seed.
RGB frames and point trajectories are aligned using
identical temporal indices.
Intervals longer than 49 frames are uniformly subsampled,
whereas intervals containing 8--48 frames are extended
to 49 frames by repeating the final RGB frame and point
positions.
The resulting padded transitions have zero ground-truth
displacement.
The repeated RGB frames and their zero-displacement
targets are included in training supervision.

\paragraph{Quality control.}
Starting from \textbf{2.30 million} raw candidate samples,
we obtain approximately \textbf{1.44 million} object-level
training samples after format standardization, quality
control, manifest validation, and exclusion of held-out
evaluation samples.
Our quality-control procedure assesses trajectories along
four dimensions: geometric validity, temporal continuity,
observation completeness, and motion plausibility.
Geometric checks reject invalid camera parameters,
non-finite 3D coordinates, and inconsistent changes
between depth and 3D position.
Temporal checks detect abrupt changes in point position,
depth, or object extent.
Observation checks limit the proportions of out-of-frame
or unobserved trajectory points.
Motion checks identify implausible displacements,
velocities, and changes in spatial scale.

The criteria depend on the reliability of each source's
geometry and scale.
For motion magnitude, SpatialCog and Real3 retain peak
displacements in $[0.05,3]$\,m, while metric MolmoMotion-1M
subsets impose an upper limit of 3\,m.
To detect temporal discontinuities, Real3 limits the
99.5th percentile of frame-rate-normalized step displacement
to 0.5\,m; metric MolmoMotion-1M subsets additionally cap
individual steps at 1\,m and speed at 15\,m/s.
DROID also rejects concurrent discontinuities in depth
and 3D position.
For spatial consistency, Real3 and metric MolmoMotion-1M
subsets limit the increase in RMS spatial radius to 0.25\,m.
For observation completeness, these sources cap the
out-of-frame fraction at 5\%; metric MolmoMotion-1M subsets
also cap the unobserved fraction at 5\%.

For YT-VIS, which lacks a reliable metric scale, we instead
use relative step-to-range, adjacent-depth, and spatial-radius
statistics without absolute displacement or speed thresholds.
Simulation records are screened for complete geometry and
valid task annotations.
All motion and continuity checks are evaluated before
temporal padding.

\begin{figure}[t]
\centering
\includegraphics[width=\linewidth]
{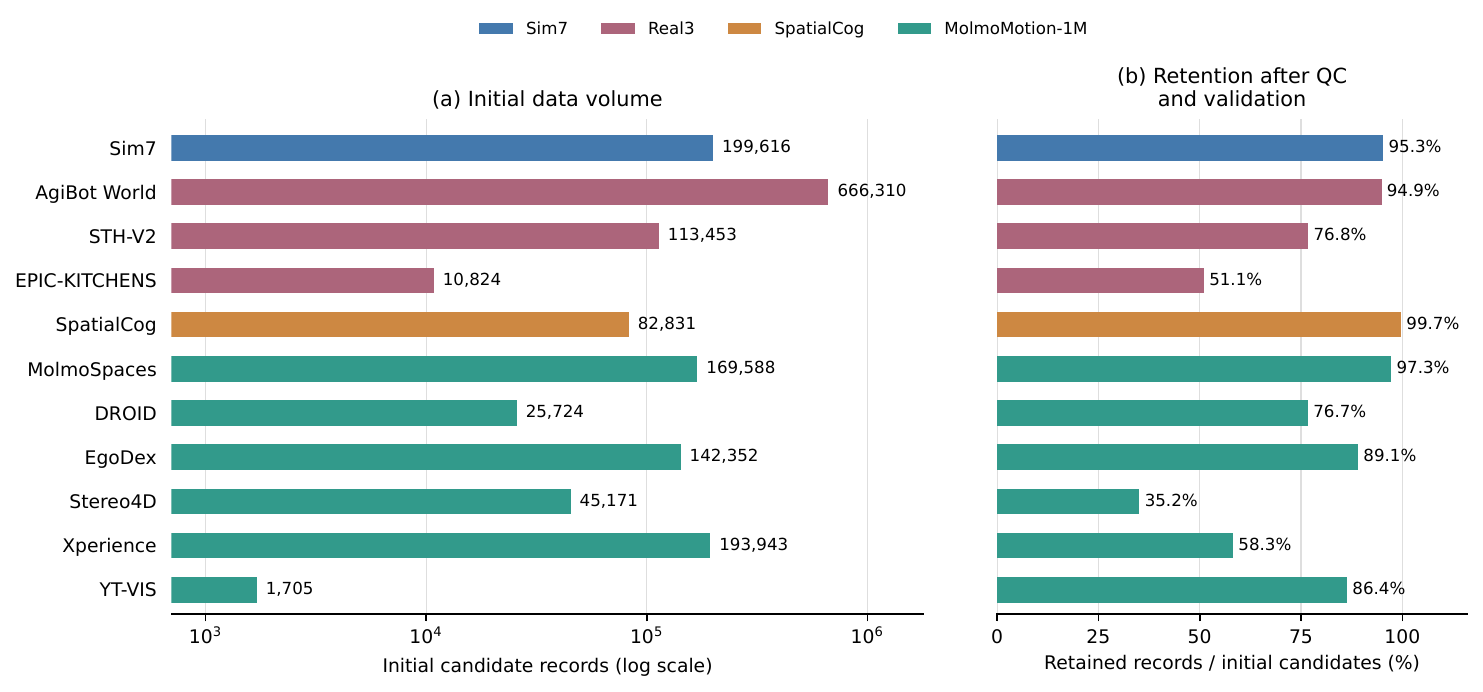}
\caption{
Initial data volume and retention across sources before
the training/evaluation split.
(a) Number of candidate object records entering the
reported processing stage, shown on a logarithmic scale.
(b) Fraction retained after formatting, quality control,
and manifest validation, before benchmark exclusion.
For SpatialCog, 82,831 candidate records yield 82,588
valid records, corresponding to a retention rate of 99.7\%.
Initial counts describe source-specific candidate pools
rather than the full original datasets.
Colors indicate the four collections defined in the main
paper.
All counts refer to object records rather than unique videos.
}
\label{fig:supp-data-composition}
\end{figure}

Figure~\ref{fig:supp-data-composition} summarizes the
source-specific candidate counts and retention after
formatting, quality control, and manifest validation,
before the training/evaluation split.
For SpatialCog, 82,831 post-prefilter candidate records
yield 82,588 valid manifest records, corresponding to
a retention rate of 99.7\%.
This pool includes all 400 SpatialCog evaluation samples
in WorldFlowBench.
These samples are excluded at training time, leaving
82,188 SpatialCog records for training.

\subsection{Dataset Statistics}

Table~\ref{tab:supp-training-data} summarizes the processed
data inventory. Real3 and SpatialCog exclude WorldFlowBench
records; Sim7 uses its QC inventory before runtime exclusion.
The reported 1,438,379 standardized object clips comprise
70,480,571 frame instances, or approximately 1,223.6 hours
at 16 FPS. With 64 query points and 48 displacement steps
per clip, they contain 92,056,256 point trajectories and
4,418,700,288 3D displacement vectors.
These totals precede Sim7 runtime exclusion and may include
padded frames, repeated query points, and multiple object
records sharing the same source video.

\begin{table}[t]
\centering
\footnotesize
\setlength{\tabcolsep}{3.5pt}
\renewcommand{\arraystretch}{1.12}
\begin{tabular}{@{}ll>{\raggedright\arraybackslash}p{0.36\linewidth}rr@{}}
\toprule
Source
& Collection
& Description
& \shortstack{Object\\clips}
& \shortstack{Standardized\\duration (h)} \\
\midrule
Sim7
& Sim7
& Simulated robot manipulation drawn from seven simulation datasets.
& 190,267 & 161.9 \\
\cmidrule(lr){1-5}

AgiBot World
& Real3
& Real-world robot manipulation across diverse environments and tasks.
& 631,873 & 537.5 \\
\cmidrule(lr){1-5}

STH-V2
& Real3
& Human--object interactions covering a broad range of object motions.
& 86,811 & 73.8 \\
\cmidrule(lr){1-5}

EPIC-KITCHENS
& Real3
& Egocentric human activities and object interactions in kitchens.
& 5,337 & 4.5 \\
\cmidrule(lr){1-5}

SpatialCog
& SpatialCog
& Tabletop and household interactions emphasizing spatial relations.
& 82,188 & 69.9 \\
\cmidrule(lr){1-5}

MolmoSpaces
& MolmoMotion-1M
& Simulated interactive scenes with released 3D motion annotations.
& 164,970 & 140.3 \\
\cmidrule(lr){1-5}

DROID
& MolmoMotion-1M
& Real-world robot manipulation collected across multiple environments
and tasks.
& 19,739 & 16.8 \\
\cmidrule(lr){1-5}

EgoDex
& MolmoMotion-1M
& Egocentric dexterous human--object interactions.
& 126,850 & 107.9 \\
\cmidrule(lr){1-5}

Stereo4D
& MolmoMotion-1M
& Dynamic real-world scenes reconstructed from stereo observations.
& 15,879 & 13.5 \\
\cmidrule(lr){1-5}

Xperience
& MolmoMotion-1M
& Egocentric videos of human activities and object interactions.
& 112,992 & 96.1 \\
\cmidrule(lr){1-5}

YT-VIS
& MolmoMotion-1M
& Diverse Internet videos with non-metric object trajectories.
& 1,473 & 1.3 \\
\midrule
Total
& ---
& ---
& 1,438,379
& 1,223.6 \\
\bottomrule
\end{tabular}
\caption{
Composition of the processed data inventory.
Real3 and SpatialCog counts exclude WorldFlowBench evaluation
records; Sim7 uses its QC inventory before runtime benchmark
exclusion.
Each object clip contains 49 frames standardized to 16 FPS.
Standardized duration in hours is computed as
$N_{\mathrm{clips}} \times 49 / (16 \times 3600)$,
where $N_{\mathrm{clips}}$ denotes the reported number of
object clips.
This measures standardized data volume rather than
deduplicated source-video duration, since different object
records may share the same underlying video interval.
}
\label{tab:supp-training-data}
\end{table}

\begin{figure}[t]
\centering
\includegraphics[width=\linewidth]
{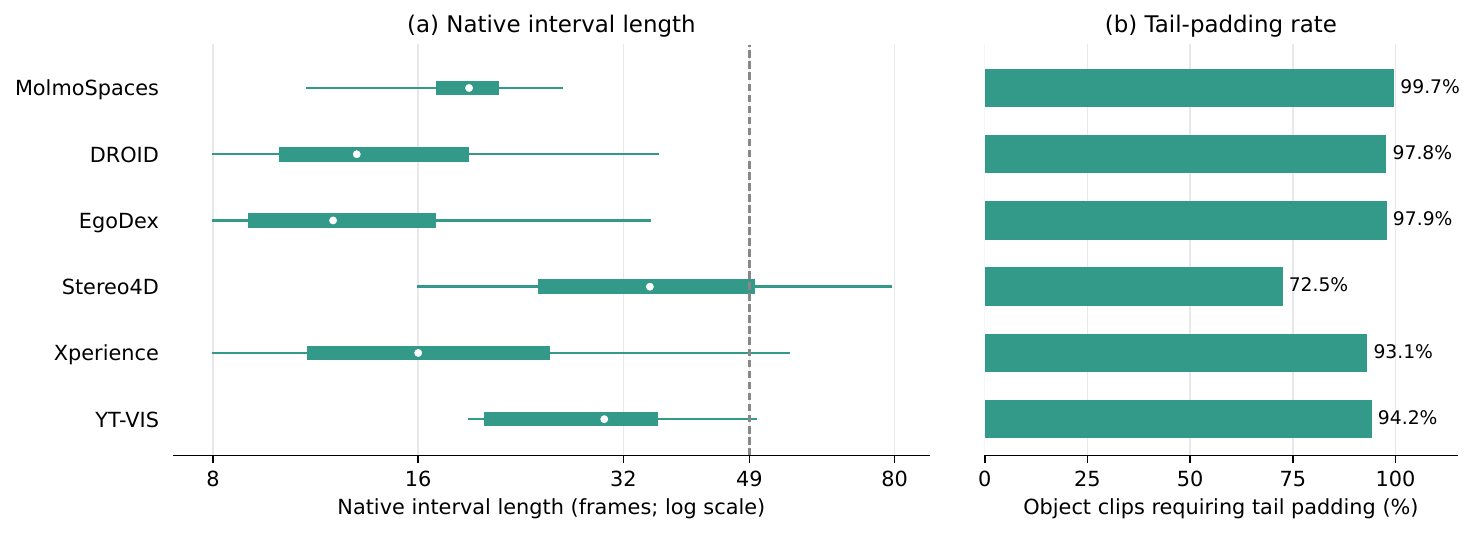}
\caption{
Native temporal coverage of the six retained MolmoMotion-1M
subsets before conversion to 49 frames.
(a) Native interval-length distributions; thin lines
indicate P5--P95, thick bars indicate the interquartile
range, and circles indicate the median.
The dashed line marks the target length of 49 frames.
(b) Percentage of records requiring tail padding.
Intervals longer than 49 frames are uniformly subsampled.
}
\label{fig:supp-native-length}
\end{figure}

The MolmoMotion-1M sources exhibit substantial variation
in their native temporal coverage.
Figure~\ref{fig:supp-native-length} reports their interval
lengths before conversion to the common 49-frame
representation and the fraction of records requiring
tail padding.

\begin{figure}[t]
\centering
\includegraphics[width=\linewidth]
{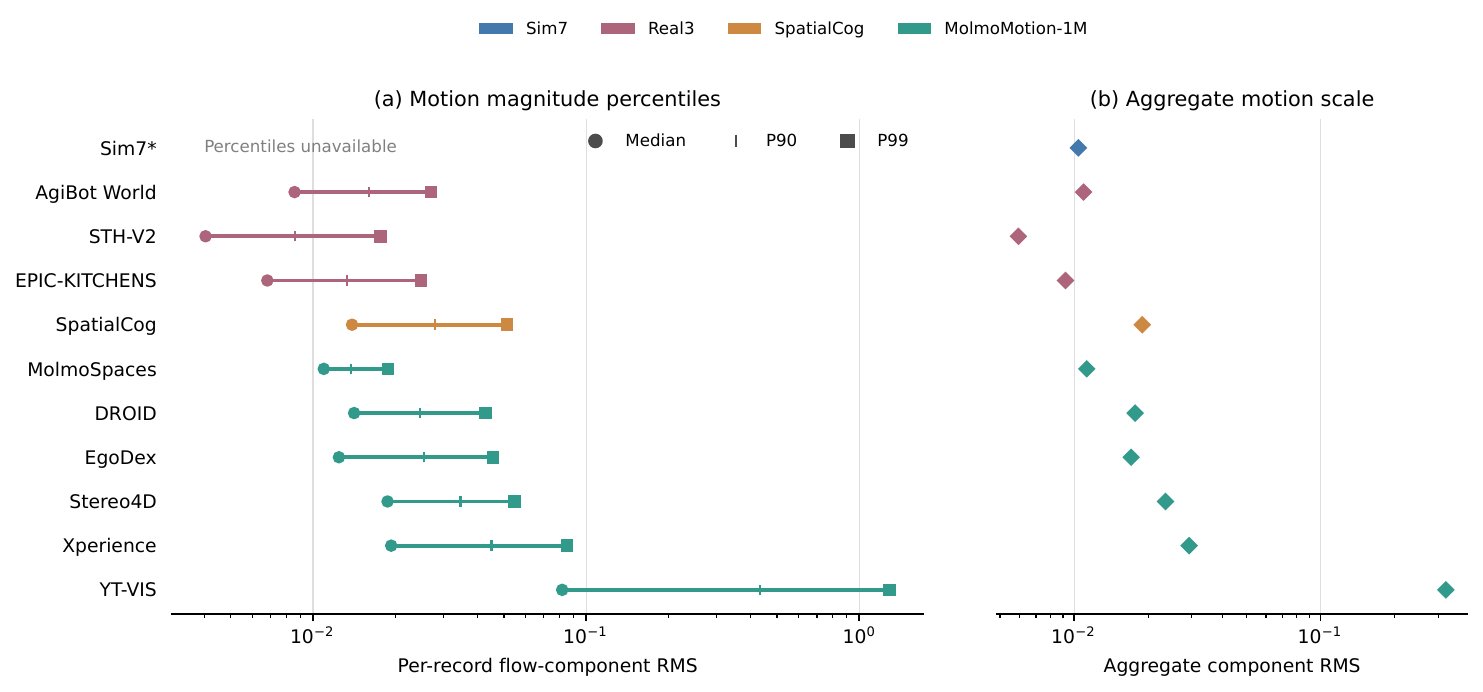}
\caption{
Pre-normalization motion-scale statistics.
Left: median, 90th percentile, and 99th percentile of
the per-record component RMS.
Right: source-level aggregate component RMS.
Both horizontal axes use logarithmic scales.
SpatialCog statistics use 82,188 records after excluding
all 400 evaluation samples; Real3 statistics also exclude
benchmark records.
Sim7, marked with an asterisk, uses its QC inventory before
runtime benchmark exclusion and provides aggregate moments
but no per-record percentile summary.
Statistics include zero-motion transitions introduced by
temporal padding.
YT-VIS values follow the non-metric scale of its annotations.
}
\label{fig:supp-motion-statistics}
\end{figure}

To characterize the scale of the motion targets, we define
the component RMS of record $r$ as
\begin{equation}
m_r
=
\left(
\frac{1}{48\cdot64\cdot3}
\sum_{t=1}^{48}
\sum_{i=1}^{64}
\left\lVert\mathbf{f}_{r,i,t}\right\rVert_2^2
\right)^{1/2}.
\label{eq:supp-motion-rms}
\end{equation}
Figure~\ref{fig:supp-motion-statistics} reports the
distribution of $m_r$ within each source together with
the source-level aggregate RMS,
$\sqrt{\mathbb{E}_r[m_r^2]}$.
These statistics are computed before flow normalization
and include zero-motion transitions introduced by temporal
padding.
They characterize the numerical scale of the motion targets
rather than physical speed or endpoint displacement.

\section{Training Details}
\label{sec:supp-training}

\subsection{Data Sampling and Curriculum}

The second stage is initialized from the EMA weights of the SpatialCog
warm-start checkpoint, with optimizer and scheduler states reset.
We detail the source-sampling strategy, curriculum transitions, structured
text conditioning, and learning-rate schedule used during its 51,100
optimization steps.

\paragraph{Hierarchical source sampling.}
We organize the training sources into five sampling domains.
Simulation combines Sim7 and MolmoSpaces; Molmo-real comprises DROID,
EgoDex, Stereo4D, and Xperience; Outdoor contains YT-VIS; and Real3 and
SpatialCog retain their collection-level definitions.
These groups determine curriculum sampling and differ from the four
data collections described in the main paper.

At optimization step $k$, the probability of selecting source $s$
in domain $d$ is
\begin{equation}
p_k(s)
=
p_k(d)
\frac{N_s^{\alpha_d}}
{\sum_{s' \in d} N_{s'}^{\alpha_d}},
\qquad
\alpha_d =
\begin{cases}
0.8, & d = \mathrm{Real3}, \\
0.5, & \text{otherwise}.
\end{cases}
\label{eq:supp-source-sampling}
\end{equation}
Here, $p_k(d)$ is the curriculum-dependent domain probability, and $N_s$
is the source count used to compute the sampler's within-domain
probabilities.
The sublinear exponent reduces the dominance of larger sources within
each domain. Sim7 is treated as a single source and retains the natural
proportions of its constituent datasets.

\paragraph{Curriculum intervals.}
Table~\ref{tab:supp-curriculum} specifies five sampling intervals that
implement the three phases described in the main paper: simulation-focused
training, mixed real-world adaptation, and SpatialCog refinement.
Real-world adaptation first emphasizes Molmo-real, briefly introduces
Outdoor data, and then prioritizes Real3. Each interval retains
cross-domain replay.

We linearly interpolate adjacent domain-probability vectors over windows
centered at steps 12,000, 24,000, 24,100, and 48,100, with widths of
1,000, 50, 50, and 1,000 steps, respectively.
Outside these transition windows, domain probabilities remain fixed at
the values in Table~\ref{tab:supp-curriculum}.
With a global batch size of 128, the nominal second-stage budget is
6,540,800 sample draws, including repeated visits to the same clips.

\begin{table}[t]
\centering
\scriptsize
\setlength{\tabcolsep}{3pt}
\renewcommand{\arraystretch}{1.1}
\begin{tabular*}{\linewidth}
{@{\extracolsep{\fill}}lrrrrrr@{}}
\toprule
Interval
& Steps
& Simulation
& Molmo-real
& Outdoor
& Real3
& SpatialCog \\
\midrule
Simulation-focused
& $[0,12{,}000)$
& 80\% & 5\% & 0\% & 10\% & 5\% \\

Molmo-real adaptation
& $[12{,}000,24{,}000)$
& 10\% & 70\% & 0\% & 15\% & 5\% \\

Outdoor adaptation
& $[24{,}000,24{,}100)$
& 5\% & 20\% & 50\% & 25\% & 0\% \\

Real3 adaptation
& $[24{,}100,48{,}100)$
& 10\% & 10\% & 0\% & 75\% & 5\% \\

SpatialCog refinement
& $[48{,}100,51{,}100)$
& 5\% & 10\% & 0\% & 35\% & 50\% \\
\bottomrule
\end{tabular*}
\caption{Target domain-sampling probabilities for the second-stage
curriculum. Adjacent distributions are linearly interpolated near interval
boundaries using the windows specified in the text. Within each domain,
sources are sampled according to Eq.~\eqref{eq:supp-source-sampling}.}
\label{tab:supp-curriculum}
\end{table}

\paragraph{Structured text conditioning.}
Each prompt consists of the original task instruction followed by a
context block with four categorical fields.
Data origin distinguishes simulated from real-world observations.
Actor type is specified as robot, human, mixed, or none.
Camera context distinguishes egocentric, fixed external, and moving
external views. Environment is specified as indoor, outdoor, or mixed.
These attributes provide explicit context for interpreting the requested
motion across heterogeneous sources.

We use sample-level context metadata when available and predefined
source-level defaults otherwise. The context fields follow a fixed order:
origin, actor, view, and scene. Training and inference use the same prompt
format, which excludes dataset identifiers and flow-normalization values.

\begin{figure}[t]
\centering
\includegraphics[width=\linewidth]
{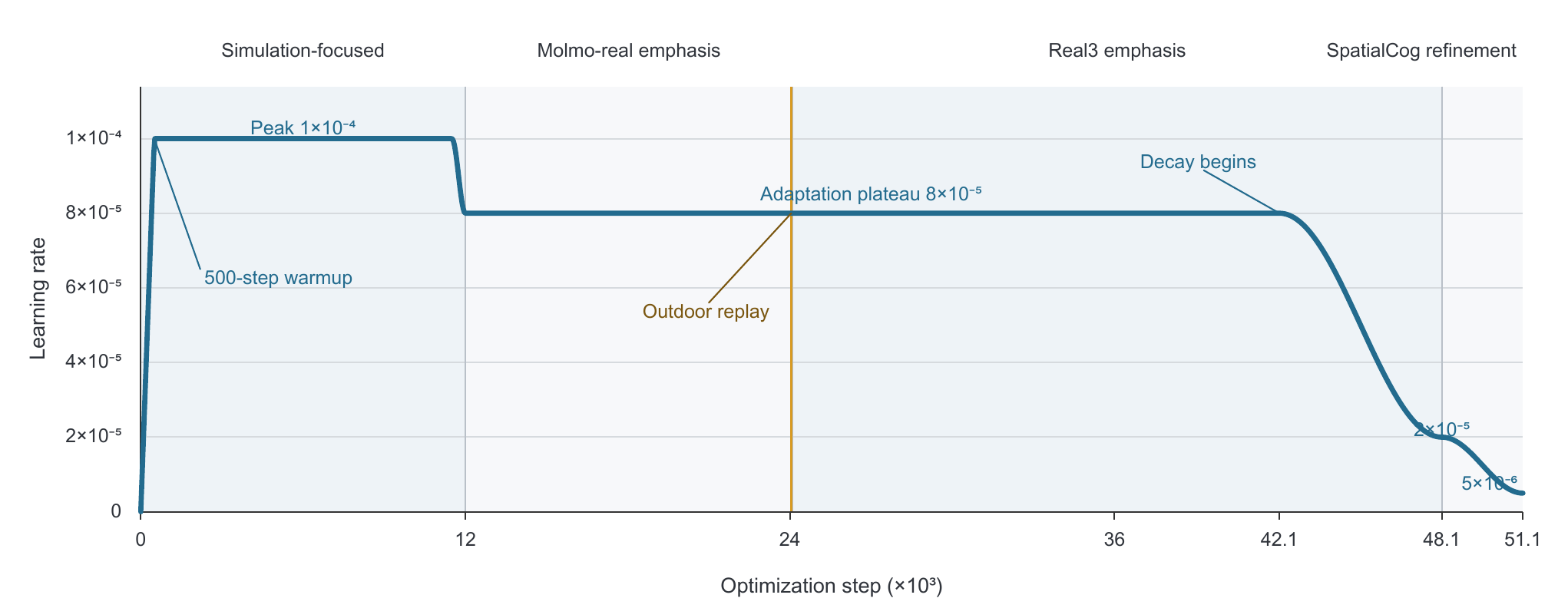}
\caption{Learning-rate schedule for second-stage training.
Background bands indicate the curriculum intervals. The narrow gold band
marks the 100-step Outdoor interval, during which the learning rate remains
at $8\times10^{-5}$.}
\label{fig:supp-curriculum}
\end{figure}

\paragraph{Learning-rate schedule.}
Figure~\ref{fig:supp-curriculum} shows the second-stage learning-rate
schedule. The learning rate increases linearly to $10^{-4}$ over the first
500 steps and remains at this level during the main simulation-focused
interval. Near the transition to real-world adaptation, it decreases to
$8\times10^{-5}$ and remains at this plateau until step 42,100, including
the brief Outdoor interval. It then decays to $2\times10^{-5}$ at step
48,100 and reaches $5\times10^{-6}$ at the end of SpatialCog refinement.

\section{Auxiliary Motion Objectives}
\label{sec:supp-aux}

The auxiliary objectives supervise frame-to-frame displacement,
cumulative displacement direction, and final displacement,
all in normalized coordinates.

\subsection{Clean-Flow Reconstruction}

We reconstruct clean displacements from the predicted rectified-flow
vector field for auxiliary supervision. Let $\mathbf{x}_0$ denote the
noise-free displacement sequence and $\sigma\in[0,1]$ the noise level.
Under the linear interpolation
$\mathbf{x}_{\sigma}=(1-\sigma)\mathbf{x}_0+\sigma\boldsymbol{\epsilon}$,
the target vector field is
$\mathbf{u}=\boldsymbol{\epsilon}-\mathbf{x}_0$.
The clean-flow estimate is
\begin{equation}
\widehat{\mathbf{x}}_0
=\mathbf{x}_{\sigma}-\sigma\mathbf{v}_{\theta}.
\label{eq:supp-clean-reconstruction}
\end{equation}
The network retains its vector-field parameterization;
auxiliary losses act on the reconstructed sequence.
Here, $x_0$ denotes the noise-free sequence, not the first video frame.

Substituting $\mathbf{x}_{\sigma}=\mathbf{x}_0+\sigma\mathbf{u}$
into Eq.~\eqref{eq:supp-clean-reconstruction} gives
$\widehat{\mathbf{x}}_0-\mathbf{x}_0
=-\sigma(\mathbf{v}_{\theta}-\mathbf{u})$, and hence
\begin{equation}
\|\widehat{\mathbf{x}}_0-\mathbf{x}_0\|_F^2
=\sigma^2\|\mathbf{v}_{\theta}-\mathbf{u}\|_F^2.
\label{eq:supp-clean-error}
\end{equation}
Thus, squared displacement supervision is exactly a noise-dependent
reweighting of the vector-field squared error.

\subsection{Motion Supervision}

Let $\mathbf{f}_{ti}$ and $\widehat{\mathbf{f}}_{ti}$ denote the
ground-truth and reconstructed normalized displacements of point $i$
at step $t$. Their cumulative displacements from the initial position
are $\mathbf{d}_{ti}=\sum_{\tau=1}^{t}\mathbf{f}_{\tau i}$ and
$\widehat{\mathbf{d}}_{ti}
=\sum_{\tau=1}^{t}\widehat{\mathbf{f}}_{\tau i}$.
Each sequence contains $T=48$ displacement steps and $N=64$ points.
We omit the batch index below for readability.

\paragraph{Adjacent-displacement loss.}
The displacement loss penalizes the squared difference between predicted
and ground-truth motion at each step:
\begin{equation}
\ell_{\mathrm{vmag}}(t,i)
=\|\widehat{\mathbf{f}}_{ti}-\mathbf{f}_{ti}\|_2^2.
\label{eq:supp-vmag}
\end{equation}
This loss supervises both magnitude and direction of local 3D
displacements, with its relation to vector-field supervision given
by Eq.~\eqref{eq:supp-clean-error}.

\paragraph{Cumulative-direction loss.}
Small local errors can accumulate into an incorrect overall direction.
We therefore compare the directions of the cumulative displacements:
\begin{equation}
\ell_{\mathrm{dir}}(t,i)
=1-\operatorname{cos}_{\varepsilon}
(\widehat{\mathbf{d}}_{ti},\mathbf{d}_{ti}).
\label{eq:supp-dir}
\end{equation}
Each error is weighted by the ground-truth magnitude
$\|\mathbf{d}_{ti}\|_2$, reducing the contribution of points with little
net displacement. For numerical stability,
$\operatorname{cos}_{\varepsilon}$ computes the dot product divided by
the product of the two vector norms, with each norm clamped below at
$\varepsilon=10^{-6}$.

\paragraph{Relative endpoint loss.}
To constrain the final predicted position, we measure endpoint error
relative to the ground-truth net displacement:
\begin{equation}
\ell_{\mathrm{end}}(i)
=\frac{
\|\widehat{\mathbf{d}}_{Ti}-\mathbf{d}_{Ti}\|_2
}{\max(\|\mathbf{d}_{Ti}\|_2,\delta)},
\qquad \delta=0.1.
\label{eq:supp-end}
\end{equation}
Only points satisfying $\|\mathbf{d}_{Ti}\|_2\geq\delta$ contribute.
This excludes small-net-displacement trajectories, including stationary
motion and motion that returns close to its starting position, for which
relative endpoint error is unstable. The threshold is defined in normalized
coordinates.

\paragraph{Weighting and reduction.}
For sample $b$, all three objectives use the noise weight
$w_b=1-\sigma_b$. The displacement loss averages
$\ell_{\mathrm{vmag}}$ over samples, steps, and points with weights $w_b$.
The direction loss averages $\ell_{\mathrm{dir}}$ over the same indices
with weights $w_b\|\mathbf{d}_{bti}\|_2$.
The endpoint loss averages $\ell_{\mathrm{end}}$ over eligible sample--point
pairs with weights $w_b$. Each weighted sum is divided by its total weight
plus $10^{-6}$; if no endpoint is eligible, its loss is zero.

This weighting gives high-noise samples relatively less influence within
each batch. For the squared displacement term, the coefficient on
vector-field squared error before batch normalization is
$\sigma^2(1-\sigma)$, by Eq.~\eqref{eq:supp-clean-error}.
The direction and endpoint terms impose additional trajectory-level
constraints; they do not share the squared-error equivalence.
Their effects are evaluated through the ablation experiments in the
main paper.

\paragraph{Training configuration.}
For second-stage training, the reduced losses use
$\lambda_{\mathrm{dir}}=0.1$,
$\lambda_{\mathrm{vmag}}=0.1$, and
$\lambda_{\mathrm{end}}=0.05$.
\section{Video-Motion Consistency}
\label{sec:joint_generation_evaluation}
\label{sec:consistency}

We evaluate whether jointly generated videos and 3D trajectories
describe consistent motion by comparing video-tracked query points
with the corresponding projected 3D predictions.
We report pixel-space discrepancies, sensitivity to tracking
confidence, and qualitative correspondence across future frames.

\paragraph{Setup.}
We evaluate video-motion consistency on all 1,678 WorldFlowBench
test clips (396 Sim7, 882 Real3, and 400 SpatialCog).
For each input, \method{} jointly generates a video and 64 query-point
trajectories in a single sampling call, covering 48 future frames
at $480\times288$ resolution.
We use Track4World~\citep{lu2026track4world} to track the initial
query pixels in the generated video without access to the predicted
flow, and Depth Anything 3~\citep{depthanything3} to estimate depth
and camera poses. We express the poses relative to the first frame
and calibrate their translation scale using the median ratio of
input query depths to estimated first-frame depths.
The predicted 3D trajectories are then projected using these poses
and first-frame intrinsics. No ground-truth future geometry is used.
All pixel distances are measured at $480\times288$ resolution.

\paragraph{Video-motion consistency metric.}
For clip $s$ and frame $t$, we retain finite, in-frame tracks whose
visibility-confidence product exceeds $\tau$, forming the query
set $\mathcal{V}_{s,t}$. We measure the mean distance between
tracked positions $\mathbf{v}_{s,i,t}$ and corresponding projected
predictions $\mathbf{u}_{s,i,t}$:
\begin{equation}
d_s(t)=\frac{1}{|\mathcal{V}_{s,t}|}
\sum_{i\in\mathcal{V}_{s,t}}
\left\|\mathbf{u}_{s,i,t}-\mathbf{v}_{s,i,t}\right\|_2.
\label{eq:consistency}
\end{equation}
Filtering depends only on tracking: out-of-frame flow projections
remain included, and nonprojectable predictions receive an
image-diagonal penalty. We average clip-level distances equally
over clips with eligible observations. At each frame, the retained
proportion is computed against all query points across all clips.

\begin{figure}[t]
\centering
\includegraphics[width=\linewidth]
{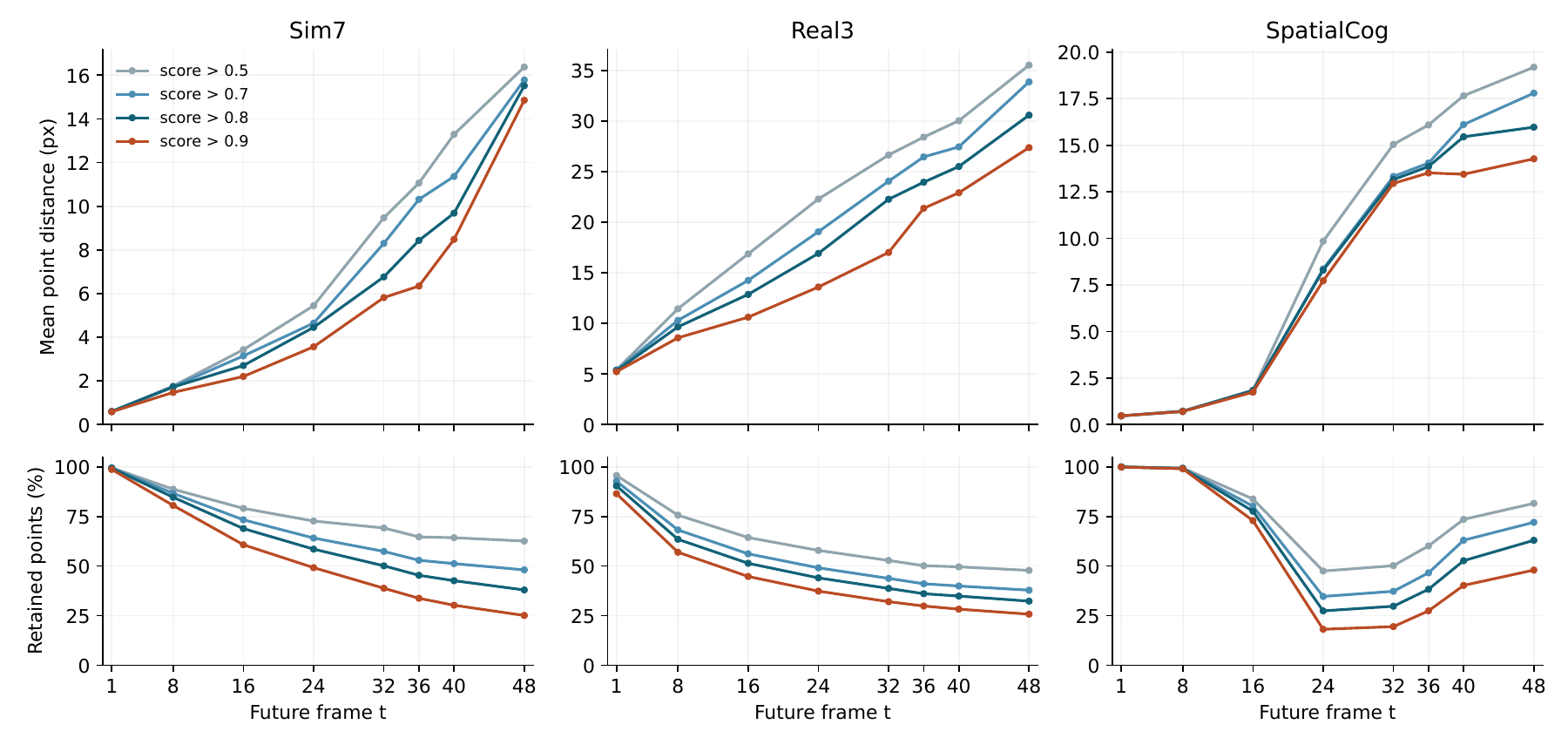}
\caption{
\textbf{Video-motion consistency under tracking-confidence filtering.}
Top: mean pointwise pixel distance at $480\times288$ resolution.
Bottom: the proportion of query points retained at each frame.
Columns correspond to Sim7, Real3, and SpatialCog.
Thresholds apply to the tracker visibility-confidence product.
Generated videos, 3D predictions, and camera projections remain
fixed across thresholds.
}
\label{fig:consistency_confidence_sweep}
\end{figure}

\paragraph{Results and confidence sensitivity.}
\method{} exhibits smaller video-motion discrepancies on tracks
assigned higher confidence by the tracker.
As shown in Figure~\ref{fig:consistency_confidence_sweep},
increasing the tracking threshold reduces the measured discrepancy
in all three domains. Across WorldFlowBench, raising $\tau$ from
0.5 to 0.9 decreases the mean pointwise distance over
48 future frames from 12.30 to 7.31 pixels.
This mean averages eligible query-frame distances within each
clip, then equally across supported clips.
These results indicate closer correspondence between generated
video and 3D motion on higher-confidence tracked observations.
The figure also reports the proportion of observations retained
at each threshold. Generated outputs and camera projections
remain fixed; only the retained query subset changes.
The measured discrepancies include residual tracking and
camera-estimation errors.

\begin{figure}[t]
\centering
\includegraphics[width=\linewidth]
{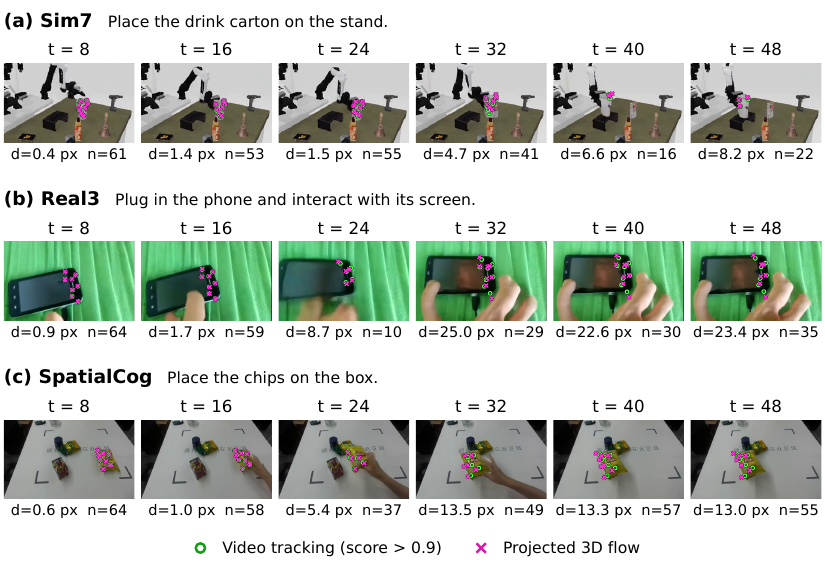}
\caption{
\textbf{Video-motion consistency across future frames.}
Rows show Sim7, Real3, and SpatialCog examples at frames
8, 16, 24, 32, 40, and 48.
Green circles and magenta crosses mark tracked video points
and corresponding projected 3D predictions, respectively.
Up to eight fixed queries with tracking score $>0.9$ are shown;
$d$ and $n$ report mean pixel distance and count over all
eligible queries.
Sim7 is cropped for display; distances use $480\times288$ coordinates.
}
\label{fig:consistency_examples}
\end{figure}

Discrepancies generally increase with the prediction horizon.
At $\tau=0.9$, frame-48 distances are 14.85, 27.39, and
14.27 pixels on Sim7, Real3, and SpatialCog, respectively,
at $480\times288$ resolution.
Figure~\ref{fig:consistency_examples} complements these measurements
by overlaying video-tracked points and corresponding projected
3D predictions at the same frame, using the same $>0.9$ threshold
and displaying no trajectory tails.
The examples show RoboTwin manipulation (Sim7), a moving phone
(Real3), and tabletop object transport (SpatialCog), illustrating
the spatial correspondence and residual discrepancies between
the two generated modalities over time.

\end{document}